\documentclass{Interspeech2024}

\interspeechcameraready

\title{Empowering Low-Resource Language ASR via Large-Scale Pseudo Labeling}

\name[affiliation={1,2}]{Kaushal}{Santosh Bhogale}
\name[affiliation={2}]{Deovrat}{Mehendale}
\name[affiliation={2}]{Niharika}{Parasa}
\name[affiliation={1,2}]{Sathish}{Kumar Reddy G}
\name[affiliation={1,2}]{Tahir}{Javed}
\name[affiliation={3}]{Pratyush}{Kumar}
\name[affiliation={1,2}]{Mitesh}{M. Khapra}

\address{
  $^1$Indian Institute of Technology Madras, India
  $^2$AI4Bharat, India
  $^3$Sarvam AI, India}
\email{cs22d006@cse.iitm.ac.in, deovrat.mehendale@gmail.com,
niharikasri.parasa@iiitb.ac.in,
\{gsathish8421,tahirjmakhdoomi\}@gmail.com,
\{pratyush,miteshk\}@cse.iitm.ac.in
}


\newcommand{\lab}[0]{PN-lab}
\newcommand{\unlab}[0]{PN-unlab}
\newcommand{\pseudolab}[0]{PN-pseudolab}
\newcommand{\vistaar}[0]{\textsc{Vistaar}}
\newcommand{\iv}[0]{\textsc{IndicVoices}}
\newcommand{\spring}[0]{\textsc{Spring-INX}}
\newcommand{\framework}[0]{Pratinidhi}
\newcommand{\yt}[0]{\textsc{IndicYT}}

\usepackage{parskip}

\begin{document}

\maketitle

\begin{abstract}

    In this study, we tackle the challenge of limited labeled data for low-resource languages in ASR, focusing on Hindi. Specifically, we explore pseudo-labeling, by proposing a generic framework combining multiple ideas from existing works. Our framework integrates multiple base models for transcription and evaluators for assessing audio-transcript pairs, resulting in robust pseudo-labeling for low resource languages. We validate our approach with a new benchmark, \yt{}, comprising diverse YouTube audio files from multiple content categories. Our findings show that augmenting pseudo labeled data from YouTube with existing training data leads to significant performance improvements on \yt{}, without affecting performance on out-of-domain benchmarks, demonstrating the efficacy of pseudo-labeled data in enhancing ASR capabilities for low-resource languages. The benchmark, code and models developed as a part of this work will be made publicly available.
    
\end{abstract}

\section{Introduction}

Recent advancements in ASR have lead to significant improvements for English, with single digit WERs on multiple benchmarks \cite{meta-mms,google-usm,whisper}. These English models are typically trained on a large amount of labelled data (at least 1000 hours \cite{librispeech}). However, such large scale high quality datasets are not available for low resource languages, such as, Hindi. While there are recent efforts in building labelled datasets for Indian languages, these efforts are still nascent with the amount of labelled data much lesser than that used for training large-scale English models. One alternative to overcome this challenge is to use weakly supervised data by aligning transcripts and audio files  available on public platforms like YouTube \cite{whisper, shrutilipi}. However, for low resource languages such data is again very sparse. Another alternative is to use pseudo-labeling wherein an existing high quality ASR model is used to transcribe readily available raw audio, with the resulting audio-transcript pairs added to the original training data. 

In this work, we study pseudo labeling in the context of a low resource language, viz., Hindi. While doing so, we propose a generic framework which combines ideas from existing works \cite{wang2020unsupervised,9643094,cascante2021curriculum,park2020improved,xu2020iterative,likhomanenko2020slimipl,lee2013pseudo} to provide an effective way of generating pseudo labeled data for low resource languages where a high quality base model may not be available to begin with. Our framework allows for multiple base models as pseudo transcribers so that an agreement between these weak transcribers can be used to improve the robustness of the generated pseudo labelled data. We further allow for evaluators which can assign a score to the generated audio-transcript pairs thereby allowing a mechanism to reject pairs with scores below a certain threshold. We acknowledge multiple such evaluators used in existing works \cite{10022960, duquenne2023sonar} and consider two types of evaluators in our work: one which relies on confidence scores of ASR models and another which relies on similarity scores computed using robust multimodal embeddings such as SONAR \cite{duquenne2023sonar} which map audio and transcripts in a common multimodal embedding space.  Our experiments show that while individual pseudo-transcribers and evaluators may not be adequate, using a combination of them works well in low resource settings.

Next, to evaluate the effectiveness of this framework, we create a robust benchmark called \yt{} by collecting audio files from YouTube which are available with a permissible CC-BY-4.0 license. These audio files are sourced from multiple channels corresponding to a diverse set of domains such as business, education, health, science, sports, cooking and so on. We show that starting with a base model trained on multiple data sources for Hindi, we can significantly improve the performance on \yt{} by augmenting the original training data with pseudo labelled data generated using our framework. These performance gains are consistent across all domains in \yt{}. Further, we also show that adding data from one source/genre (i.e., YouTube) does not affect the performance on out-of-domain benchmarks. To do so we evaluate the model on the comprehensive \vistaar{} \cite{vistaar} benchmark which contains test sets from multiple sources/domains. Our main contributions can thus be summarised as:
\begin{itemize}
    \item A robust framework for pseudo labeling for low resource languages
    \item A new diverse benchmark, \yt{}, for evaluating ASR models for Hindi on multiple content categories from YouTube
    \item A state of the art ASR model for Hindi, which performs well on \yt{} as well as existing benchmarks. 
\end{itemize}

\section{Related Work}

Several prior works \cite{wang2020unsupervised,9643094,cascante2021curriculum} have used pseudo-labeling as a semi-supervised learning technique for improving ASR systems. These methods typically use a combination of the following stages (i) generate pseudo labels (ii) refine/filter the pseudo-labeled data (iii) retrain the model and (iv) iterate. For example, Noisy Student Training \cite{park2020improved} filters the augmented pseudo-labeled data using a Language model (LM) between self-training iterations.  Iterative Pseudo-labeling (IPL) \cite{xu2020iterative} performs multiple iterations of pseudo-labeling on the unlabeled data as the acoustic model evolves. SlimIPL \cite{likhomanenko2020slimipl} improves upon the IPL algorithm by iteratively regenerating transcriptions with the most probable tokens (called hard labels) without a language model. \cite{lee2013pseudo}  propose a self-training approach that combines multiple models during training to improve label diversity and keep the model from being overly confident to noisy pseudo-labels.

Pseudo-labeling has also been extended to multilingual ASR \cite{lugosch2022pseudo}, by (i) training a supervised multilingual model, (ii) performing monolingual finetuning to get language specific models (iii) generate pseudo-labels using monolingual models and  then (iv)train a final multilingual model using pseudo-labels from all languages.
To improve quality of pseudo-labeling on non target domain data, \cite{chen2023improving} propose a data selection strategy using an LM, while \cite{nandi2023pseudo} uses two expert models and metrics based on confidence score and min-max audio duration to filter data.
In this work, we combine ideas from prior work and propose a generic framework for pseudo-labeling.

\section{\framework{}: A generic framework for large-scale pseudo-labeling}
In this section, we present \framework\footnote{\framework{} means representative in Sanskrit, as in, a representative of the true transcript}, a generic framework to perform pseudo-labeling on large-scale datasets by combining ideas from several existing works. 

\subsection{The Framework}
The framework consists of Pseudo-Transcribers and Evaluators, which participate in labeling and filtering, respectively. Given access to $N$ Pseudo-Transcribers and $K$ Evaluators, the framework tries to estimate the true transcript $t^*$ of an unlabeled audio segment $X$. We next describe the components of our framework.\\
\textbf{Pseudo-Transcribers.} Pseudo-transcribers are trained ASR models that can act as base models to generate pseudo-transcripts $t$ for an unlabeled audio segment. We propose using multiple pseudo-transcribers to improve quality of pseudo-labels. In practice, multiple such pseudo-transcribers can be obtained by training models with different architectures on the same dataset, or by training models on different domains, or by accessing open-source models or public speech-to-text APIs.\\
\textbf{Evaluators.} Evaluators are metrics that assign a score to pseudo-transcripts. Specifically, an Evaluator $g(\cdot)$ estimates a score $s = g(t,X)$ where $t$ is a pseudo-transcript for an unalabeled audio segment, $X$. We identify two types of metrics that can be used as evaluators: one which relies on confidence scores of ASR models \cite{10022960} and another which relies on similarity scores computed using multimodal audio/sentence embeddings \cite{duquenne2023sonar}.\\
\textbf{Pseudo-Labeling.} Given a set of $N$ psuedo-transcripts $\{t_1, t_2, .., t_N\}$, the pseudo-labeling algorithm returns $t_{max}$, the candidate that has maximum agreement with other pseudo-transcript candidates. We first calculate a text matching score between two texts $p$ and $q$, given by
\begin{equation}
    M(p,q) = 1-\dfrac{LD(p,q)}{|p|+|q|}
\end{equation}
where $LD$ is the Levenshtein distance between the
two texts and $|.|$ denotes the length of the text, in characters.
Then, for each $t_i$, we define Agreement -
\begin{equation}
    A_i^{k} =
    \begin{cases}
      1, & \text{if}\ M(t_i, t_k) \ge \tau \\
      0, & \text{otherwise}
    \end{cases}
\end{equation}
where $\tau$ is a hypermarameter.
Next, we calculate the agreement score $C_i = \sum_{k=1}^{N} A_i^n$. Finally, the pseudo-transcript $t_{acpt}$ is considered as accepted if $C_i > \delta $. If multiple pseudo-transcripts satisfy the threshold, then the transcript with the maximum $C_i$ is selected, with ties broken arbitrarily. If no transcript satisfies the above condition then this audio is not considered for pseudo labelling.\\
\textbf{Filtering.} After obtaining an initial pseudo-transript candidate $t_{acpt}$ for an audio segment $X$, it is evaluated by $K$ Evaluators. For each evaluator $k \in K$, we first calculate,
\begin{equation}
    B(t_{acpt},k) =
    \begin{cases}
      1, & \text{if}\ g_k(t_{acpt}, X) \ge \rho_k \\
      0, & \text{otherwise}
    \end{cases}
\end{equation}
where $\rho_k$ is a hyperparameter specific to each evaluator.
We define the filter score $F(t_{acpt}) = \sum_{k=1}^{K}B(t_{acpt},k)$. Finally, the pseudo-transcript $t_{acpt}$ is considered as a proxy for the true transcript if $F(t_{acpt}) > \lambda $, else the audio is not considered for pseudo-labelling.

\subsection{Application to a Low-resource Indian Language}
We show that our framework is particularly useful for low-resource languages. In this paper, we focus on Hindi, a language used by a population exceeding 600 million people, yet falling short in ASR resources when compared to English. 
Next, we describe the problem setting for Hindi including labeled datasets, unlabeled datasets, Pseudo-Transcribers and Evaluators. For convenience we abbreviate \framework as PN below.\\
\textbf{\lab{}.} To train ASR models that can act as Pseudo-Transcribers, we first collate all publicly available \textit{labeled} datasets for Hindi. Specifically, we extend the previous effort of \vistaar{} \cite{vistaar}, which collated 2150 hours of diverse training data for Hindi. We add two recently released Indian language datasets to their collection, viz., \iv{} \cite{indicvoices} and \spring{} \cite{spring-inx}, containing 65 hours and 316 hours of training data for Hindi, respectively. The total size of our combined training data is 2531 hours. We refer to it as PN-lab. \\
\textbf{\unlab.} We identify an unlabeled dataset available for Hindi -  Dhwani \cite{indicwav2vec}. It contains 1075 hours of audio from YouTube. In order to increase the scale of the dataset further, we collect publicly accessible videos from YouTube. We make use of the YouTube Search API to automate the process of discovering videos. The search terms were crafted carefully to ensure diversity in domains in the collected data. We follow the data preprocessing steps described in \cite{indicwav2vec}. After filtering and data preprocessing, we obtain a corpus of 28616 hours of unlabeled audio. We refer to it as PN-unlab. \\
\textbf{Pseudo-Transcribers.} To obtain base models, we train a Conformer-L \cite{conformer} model with a hybrid CTC-RNNT \cite{ctc-rnnt} decoder on PN-lab. We can derive two ASR models from this single training, one using the CTC head and another using the RNNT head of the hybrid model. We first evaluate the efficacy of these models on the Vistaar benchmark. As shown in Table \ref{tab:vistaar_hindi}, in comparison to other publicly available and open-source models like Google USM, Whisper, MMS, IndicWhisper, we obtain improvement of 1.7 point in WER, thus establishing it as a good base model. \\
\textbf{Evaluators.} We use two evaluators. The first is linearly normalized Renyi entropy-based confidence \cite{10022960} of the RNN-T model, as this was shown to be effective in our initial experiments.
The second is SONAR similarity \cite{duquenne2023sonar}, which computes cosine similarity using multimodal audio/sentence embeddings. Our choice of SONAR is motivated by the fact that SONAR already supports 14 Indian languages, and can be extended to other low-resource Indian languages.

\subsection{\pseudolab}
Using the \framework{} framework, we perform pseudo-labeling on \unlab{} using the RNN-T and CTC models as pseudo-transcribers. \\
We first perform pseudo-labeling with $\tau=1$ and $\delta=1$, which is equivalent to accepting the pseudo-label if there was an exact match in the transcripts of the two models. Next, we run the filtering algorithm with $\rho_{SONAR}$ set to 0.8 and $\rho_{RNNT}$ set to 0.7. The thresholds were chosen based on manually verifying the quality of accepted pseudo-transcripts on a set of samples. We used $\lambda=2$ indicating that samples were accepted when threshold criteria was met for both SONAR and RNN-T.\\
Starting with a set of 28616 hours of unlabeled data, we get 4862 hours after pseudo-labeling (the rest gets rejected as the transcripts of the 2 models do not match). Consequently, we get 1840 hours after performing the filtering using the two evaluators. Next, we use the combined set of \lab{} and \pseudolab, called PN, to retrain the base models and report results in Section 5. While the framework allows for multiple iterations of the above process, in this work we only perform one iteration.

\section{\yt Benchmark}
To allow for a robust evaluation of Hindi ASR models across multiple domains, we create and release a benchmark called \yt. This benchmark was created by first curating diverse audio data with CC-BY-4.0 license from YouTube and then transcribing a 2 hour subset of this data with the help of human annotators, ensuring high quality. Our process involved the following steps:

\textbf{Identification of topics for collection.} In order to maximize the diversity of the curated content, our search process began with identifying a list of categories that can be used by our experts to search for rich content on YouTube. We defined 3 major categories, viz., How-To-Do videos, Household activities, News media. The motivation for choosing these categories was to ensure the expansion of this search approach to any medium/low resource language. Once the broad categories were fixed, we created a list of subcategories which can be directly used to query YouTube for content. Table \ref{tab:data} show the curated list of domains and their sub-categories.\\
\textbf{Collection.} We recruited language experts proficient in Hindi to do this task. Each language expert was asked to find atleast 35 videos from a particular category, having CC-BY-4.0 license. The experts were instructed not to collect more than 3 videos from a single channel to ensure speaker diversity. \\
\textbf{Transcription.} Next, we used Shoonya as a platform to transcribe a two hour subset of collected data and transcribed it using the same guideline as discussed in \cite{indicvoices}. Before transcription, the videos were converted into audios and passed through Voice activity detection (VAD) to extract the voiced segments from the audios. Next, the subset selection was done by random sampling one VAD segment for each audio, resulting in a total of 2 hours of benchmark data.\\
\noindent\textbf{Statistics.} Table \ref{tab:data} shows some statistics about \yt. Specifically, it shows the number of minutes of transcribed data for each category, the number of channels from which the data was sourced and the number of utterances/segments as identified by VAD.\\
\begin{table}[htbp]
    \centering
    \footnotesize
    \caption{Data Statistics}
    \begin{tabular}{lrrr}
        \toprule
        Domain & Dur(mins) & \#Channels & \#Utt \\
        \midrule
        Business News & 10.48 & 33 & 76 \\
        Cooking & 11.84 & 76 & 138 \\
        Debates & 7.86 & 6 & 56 \\
        Education - Technology & 9.86 & 43 & 110 \\
        Headlines & 7.48 & 19 & 51 \\
        Health & 9.28 & 18 & 86 \\
        Household activities & 8.57 & 17 & 68 \\
        How-to Technology & 10.24 & 61 & 121 \\
        Interviews\_Panels & 9.53 & 24 & 98 \\
        Maths & 9.01 & 27 & 86 \\
        On-field reporting & 8.46 & 6 & 64 \\
        Science & 7.30 & 14 & 64 \\
        Social Science & 9.78 & 22 & 84 \\
        Sports News & 8.15 & 5 & 53 \\
        \bottomrule
    \end{tabular}
    \label{tab:data}
\end{table}

\begin{table*}[!t]
\centering
\footnotesize
\caption{Comparison of publicly-available models on the Hindi subset of the Vistaar benchmark}
\label{tab:vistaar_hindi}
\begingroup
\begin{tabular}{lcccccccc}
\toprule
Model         & Kathbath   & Kathbath-Hard & FLEURS & CommonVoice & IndicTTS & MUCS & Gramvaani & Average  \\
\midrule
Google STT        & 14.3 & 16.7    & 19.4   & 20.8 & 18.3     & 17.8 & 59.9      & 23.9 \\
IndicWav2vec  & 12.2 & 16.2    & 18.3   & 20.2 & 15.0     & 22.9 & 42.1      & 21.0 \\
Azure STT        & 13.6 & 15.1    & 24.3   & 14.6 & 15.2     & 15.1 & 42.3      & 20.0 \\
Nvidia-medium & 14.0 & 15.6    & 19.4   & 20.4 & 12.3     & 12.4 & 41.3      & 19.4 \\
Nvidia-large  & 12.7 & 14.2    & 15.7   & 21.2 & 12.2     & 11.8 & 42.6      & 18.6 \\
IndicWhisper  & 10.3 & 12.0    & 11.4   & 15.0 & 7.6     & 12.0 & 26.8     & 13.6 \\
\midrule 
PN-lab       & 8.8 & 10.0 & 11.3 & 11.2 & 6.8 &  10.0 & 23.7 & 11.9 \\
PN & \textbf{8.9} & \textbf{10.1} & \textbf{11.9} & \textbf{10.6} & \textbf{7.7} & \textbf{9.0} & \textbf{24.2} & \textbf{11.9} \\
\bottomrule
\end{tabular}
\endgroup
\end{table*}

\begin{figure}[!b]
    \centering
    \includegraphics[width=\linewidth]{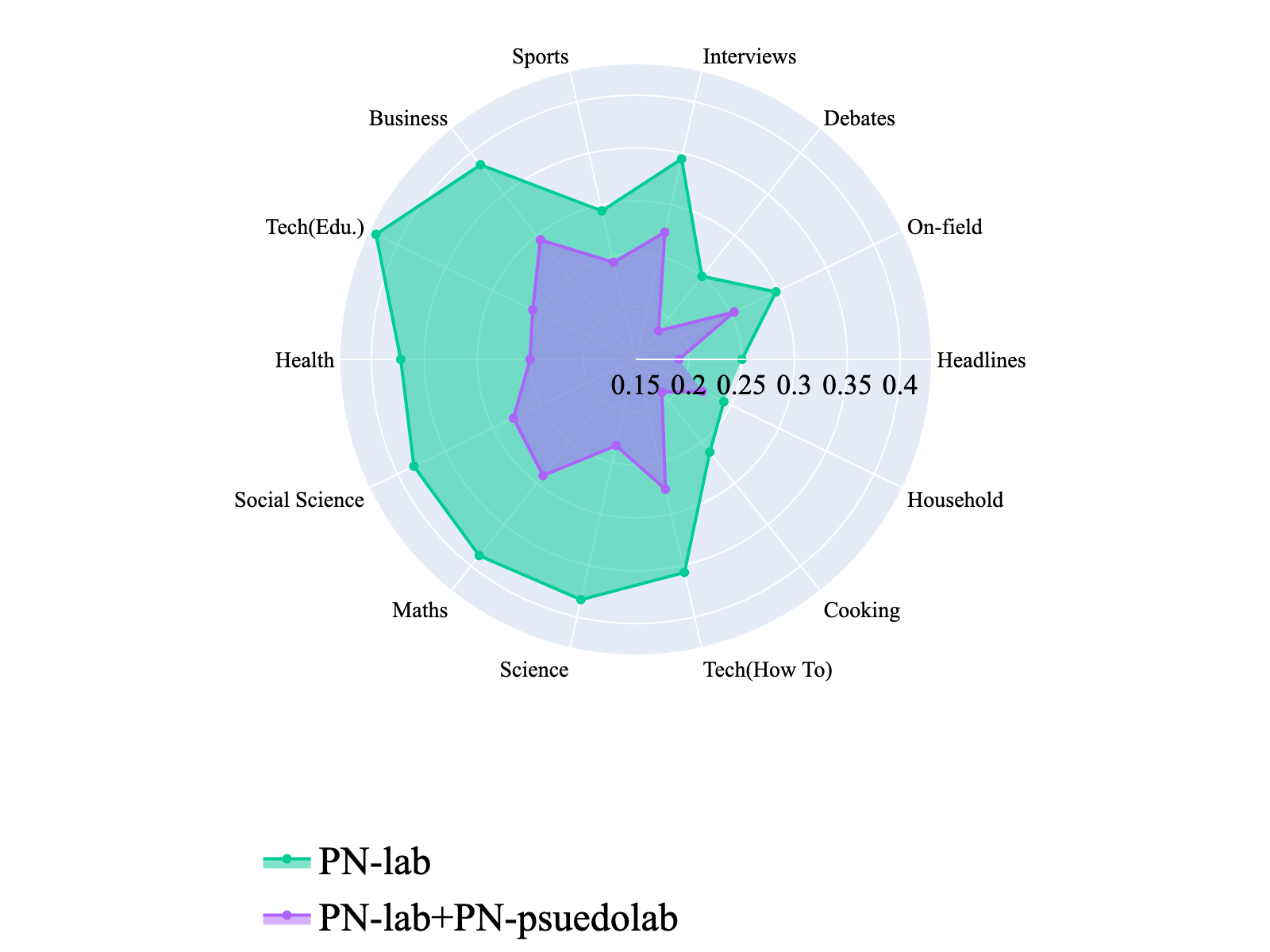}
    \caption{Comparison of our model on the \yt benchmark shows improvement across all domains, with significant improvement on Education domains like Math and Science.}
    \label{fig:domains}
\end{figure}

\section{Results and Discussion}

\textbf{Experiment Setup.}
We train Conformer-L \cite{conformer} models, consisting of 120M parameters, as the encoder, with a hybrid CTC-RNNT \cite{ctc-rnnt} decoder. The model has 17 conformer blocks with 512 as the model dimension.
All our models are trained using standard recipes from the NeMo \cite{nemo} library.
The same model is trained on \lab{} as well as \lab+\pseudolab \hspace{0.5pt} with consistent hyperparameters, to study the effect of pseudo-labelling. The results of our experiments are summarised in Tables \ref{tab:vistaar_hindi},\ref{tab:ablation},\ref{tab:comparison} and Figure \ref{fig:domains}. Below we discuss some of the interesting observations from our results. 

\noindent\textbf{Qualitative Analysis of Pseudo-Labeled Samples.} We first do a qualitative analysis of the pseudo-labeled samples generated using \framework. We randomly select 500 samples from the \pseudolab{}, spanning various domains. We then asked human evaluators to listen to audios and read the transcripts to identify errors. The evaluators reported that 325/500 transcripts had no errors. Of the 175 erroneous samples, 65 had additions, deletions or substitutions whereas remaining 110 samples had minor spelling errors, such as interchanged long and short vowels or alternative non-standard spellings for code-mixed words. 

\noindent\textbf{Comparison of our model on the \yt benchmark.}
We compare the performance of \lab{} and PN-lab+PN across the different domains of the \yt{} benchmark. We observe that the addition of the \pseudolab{} dataset helps across all domains, with an average improvement of 8.6\% across domains. Figure \ref{fig:domains} gives a holistic picture of the improvement across multiple domains. It is interesting to note that while the base model (green) performs poorly on inherently difficult domains, such as Maths, Science, Health, Tech (Edu.), adding pseudo-labeled data (purple) significantly improves the performance on these difficult domains.

\noindent\textbf{Robustness to audio lengths.} Next, we assess the robustness of the model across utterances of different lengths. Specifically, Table \ref{tab:comparison} compares the performance of the base model (Model 1) trained on \lab{} with our final model (Model 2) trained on \lab{} and \pseudolab{}, on audio segments of different lengths. We observe larger improvements for shorter and longer audios and a moderate improvement for mid-sized audios. We hypothesise that this is because existing training data largely contain medium sized audio files whereas \pseudolab{} contains short and long utterances also as the audio files are segmented using VAD which may generat shorter segments if there are frequent pauses or the speech is slow and longer segments for rapidly spoken content without notable pauses. 

\noindent\textbf{Effect of combining pseudo-transcribers and evaluators.} Next, we compare 4 different methods for pseudo-labeling (i) PN-RNNT which just uses the transcripts generated by the RNNT decoder without using any evaluator for filtering or using any other  pseudo-transcriber for matching (ii) PN-SONAR which combines the RNNT transcriber with a SONAR based evaluator to filter audio-transcript pairs which have lower cosine similarity in the SONAR embedding space (iii) PN No-Filter just relies on two pseudo transcribers (RNN-T and CTC) and does not use any evaluator based filtering and (iv) PN which is our final approach using two pseudo transcribers and the SONAR evaluator. The results in Table \ref{tab:ablation} clearly show that combining multiple pseudo-transcribers and evaluators (as in the last column, PN) gives the best results. This establishes the effectiveness of our framework integrating multiple systems for pseudo labeling for low resource languages.

\noindent\textbf{Comparison on out-of-style benchmarks.} One could argue that adding pseudo-labeled data of a specific style (YouTube) in this case, could affect performance on benchmarks containing audio which is different from YouTube content. To evaluate this, we test our final model on the robust \vistaar{} benchmark which contain audio of multiple styles, e.g., crowdsourced read speech, on-field data, call centre data and so on. The results in Table \ref{tab:vistaar_hindi} show that even though our final training data is dominated by pseudo-labeled content from YouTube, the performance on other benchmarks in not affected, with the average WER being comparable (last two rows of Table \ref{tab:vistaar_hindi}).


\begin{table}[!t]
    \small
    \centering
    \footnotesize
    \caption{Effect of combining pseudo-transribers and evaluators}
    \setlength{\tabcolsep}{2.8pt}
    \begin{tabular}{lrrrrc}
    \toprule
    \textbf{Domains}&
    \begin{tabular}[c]{@{}c@{}}\textbf{PN}\\\textbf{Lab}\end{tabular} & 
    \begin{tabular}[c]{@{}c@{}}\textbf{PN}\\\textbf{RNNT}\end{tabular}&
    \begin{tabular}[c]{@{}c@{}}\textbf{PN}\\\textbf{SONAR}\end{tabular} &
    \begin{tabular}[c]{@{}c@{}}\textbf{PN}\\\textbf{No-Filter}\end{tabular}&
    \textbf{\begin{tabular}[c]{@{}c@{}}PN \end{tabular}}\\
    \midrule
    Headlines              & 25.0            & 20.3                                 & 20.8                                  & 19.7                                   & 19.0                         \\
    On-field reporting     & 29.7            & 26.1                                 & 27.4                                  & 26.4                                   & 25.3                         \\
    Debates                & 25.0            & 23.0                                 & 20.7                                  & 20.4                                   & 18.4                         \\
    Interview - Panels     & 34.4            & 28.8                                 & 27.9                                  & 28.3                                   & 27.3                         \\
    Sports News            & 29.4            & 27.3                                 & 25.1                                  & 27.6                                   & 24.4                         \\
    Business News          & 38.5            & 33.5                                 & 30.5                                  & 30.9                                   & 29.4                         \\
    Education - Technology & 42.2            & 32.5                                 & 27.2                                  & 29.1                                   & 25.8                         \\
    Health                 & 37.2            & 35.1                                 & 26.6                                  & 25.1                                   & 25.0                         \\
    Social Science         & 38.2            & 34.6                                 & 29.2                                  & 31.4                                   & 27.8                         \\
    Maths                  & 38.7            & 32.2                                 & 29.1                                  & 28.3                                   & 29.0                         \\
    Science                & 38.3            & 34.2                                 & 26.4                                  & 27.9                                   & 23.3                         \\
    How-to Technology      & 35.6            & 30.7                                 & 28.0                                  & 27.8                                   & 27.6                         \\
    Cooking                & 26.2            & 21.1                                 & 20.1                                  & 19.5                                   & 18.9                         \\
    Household activities   & 24.2            & 21.8                                 & 22.2                                  & 22.5                                   & 21.9                         \\
    \midrule
    \textbf{ALL}           & \textbf{33.3}   & \textbf{28.8}                 & \textbf{26.0}                  & \textbf{25.0}                  & \textbf{24.7}     \\
    \bottomrule
\end{tabular}
\label{tab:ablation}
\end{table}

\section{Conclusion}
We address the challenge of limited labeled data for low-resource languages in ASR, with a specific focus on Hindi. We propose a pseudo-labeling framework
for generating high-quality pseudo-labeled data. This approach leverages the consensus among multiple base models and employs evaluators to ensure the reliability of the generated audio-transcript pairs. Our experiments reveal marked improvements in ASR performance on the \yt{} benchmark created as a part of this work, affirming the value of pseudo-labeled data in enhancing ASR capabilities for languages with scant resources, without compromising performance on external benchmarks. 

\begin{table}[t]
  \centering
  \caption{Robustness to audio lengths}
    \begin{tabular}{lccc}
    \toprule
    Duration (seconds) & PN-lab & PN \\
    \midrule
    2-10 & 35.70 & 26.29 \\
    10-20 & 28.61 & 22.57 \\
    20-30 & 34.30 & 23.27 \\
    \bottomrule
    \end{tabular}%
  \label{tab:comparison}%
\end{table}%

\section{Acknowledgements}
We would like to thank Digital India Bhashini, the Ministry of Electronics and Information Technology (MeitY\footnote{https://www.meity.gov.in/}) of the Government of India and the Centre for Development of Advanced Computing (C-DAC\footnote{https://www.cdac.in/index.aspx?id=pune}), Pune for generously supporting this work and providing us access to multiple GPU nodes on the Param Siddhi Supercomputer. We would like to thank the EkStep Foundation and Nilekani Philanthropies for their generous grant which went into hiring human resources as well as cloud resources needed for this work.

\bibliographystyle{IEEEtran}
\bibliography{mybib}

\end{document}